\documentclass[lettersize,journal]{IEEEtran}
\usepackage{amsmath,amsfonts}
\usepackage{mathabx}
\usepackage{algorithmic}
\usepackage{algorithm}
\usepackage{array}
\usepackage[caption=false,font=normalsize,labelfont=sf,textfont=sf]{subfig}
\usepackage{textcomp}
\usepackage{stfloats}
\usepackage{url}
\usepackage{verbatim}
\usepackage{graphicx}
\usepackage{cite}
\usepackage{microtype}
\usepackage{graphicx}
\usepackage{booktabs} 
\usepackage{multirow} 

\usepackage{bbding}
\usepackage{hyperref}
\usepackage{float}
\usepackage{xcolor}

\begin{document}

\title{LangSurf: Language-Embedded Surface Gaussians for 3D Scene Understanding}

\author{Hao Li*, Minghan Qin*, Zhengyu Zou*, Diqi He, Xinhao Ji, Bohan Li, Bingquan Dai, \\ Dingwen Zhang, Junwei Han\textsuperscript{\Envelope}, \textit{IEEE Fellow}
\IEEEcompsocitemizethanks{
\IEEEcompsocthanksitem Hao Li, Zhengyu Zou, Diqi He, Dingwen Zhang (Corresponding author) are with the Brain and Artificial Intelligence Lab, Northwestern Polytechnical University, Xi'an, Shaanxi 710000, China.
\IEEEcompsocthanksitem Minghan Qin (Corresponding author), Bingquan Dai are with the Tsinghua University.
\IEEEcompsocthanksitem Bohan Li is with the Shanghai Jiaotong University.
\IEEEcompsocthanksitem Xinhao Ji is with the Peking University and interns in Northwestern Polytechnical University.
\IEEEcompsocthanksitem Junwei Han is with the School of Artifical Intelligence, Chongqing University of Posts and Telecommunications.
}
\thanks{* denotes equal contributions.}
\thanks{This work was supported in part by the National Natural Science Foundation of China under Grant 62293543, Grant 62322605, and Grant 625B2148.}
}

\markboth{Journal of \LaTeX\ Class Files,~Vol.~14, No.~8, August~2021}%
{Shell \MakeLowercase{\textit{et al.}}: A Sample Article Using IEEEtran.cls for IEEE Journals}

\IEEEpubid{0000--0000/00\$00.00~\copyright~2021 IEEE}

\maketitle

\begin{abstract}
    Applying Gaussian Splatting to perception tasks for 3D scene understanding is becoming increasingly popular. 
Most existing works primarily focus on rendering 2D feature maps from novel viewpoints, which leads to an imprecise 3D language field with outlier languages, ultimately failing to align objects in 3D space.
%
%
To this end, we propose a Language-Embedded Surface Field (LangSurf), which accurately aligns the 3D language fields with the surface of objects, facilitating precise 2D and 3D segmentation with text query, widely expanding the downstream tasks such as removal and editing.
The core of LangSurf is a joint training strategy that flattens the language Gaussian on the object surfaces using geometry supervision and contrastive losses to assign accurate language features to the Gaussians of objects.
In addition, we also introduce the Hierarchical-Context Awareness Module to extract features at the image level for contextual information then perform hierarchical mask pooling using masks segmented by SAM to obtain fine-grained language features in different hierarchies.
Extensive experiments on open-vocabulary 2D and 3D semantic segmentation demonstrate that LangSurf outperforms the previous state-of-the-art method LangSplat by a large margin. As shown in Fig.~\ref{fig:teaser}, our method 
 is capable of segmenting objects in 3D space, thus boosting the effectiveness of our approach in instance recognition, removal, and editing, which is also supported by comprehensive experiments.
\end{abstract}


\section{Introduction}
    \label{sec:intro}
\IEEEPARstart{R}{ecently} , 3D scene understanding has emerged as a critical research focus. By integrating natural language with 3D scenes, systems can enable more intuitive human-computer interactions in applications such as virtual reality\cite{li2022rt, jiang2024vr,jaritz2019multi,gpnerf}, autonomous driving\cite{kitti,zou2025m}, and robotics~\cite{awais2023foundational, rnrmap, kerr2024robot, zhang2025category, firoozi2023foundation, lu2025manigaussian}. However, accurately embedding semantic information within 3D space remains a significant challenge.

Current methods use NeRF (Neural Radiance Fields)~\cite{nerf} or 3DGS (3D Gaussian splatting)~\cite{3dgs} as the 3D representation, combined with language features from the CLIP~\cite{clip} model, to enable open-vocabulary 3D querying.
However, while the semantic maps generated in these methods are critical for supervising the 3D semantic field, they lack sufficient contextual information~\cite{pspnet}. More specifically, these methods typically rely on sliding windows (as in LERF~\cite{lerf}) or Segment Anything Model (SAM~\cite{sam}) masks (as in LangSplat~\cite{langsplat}) to divide images into parts, which are then processed by CLIP to extract the corresponding semantic features. In such a process, the obtained semantic features only contain information from the local image regions, which can hardly be used to represent the semantics of low-texture regions~\cite{attention}, such as walls and floors, or complex object structures that are divided into multiple parts. 

Additionally, LERF and LangSplat primarily focus on rendering 2D feature maps from novel viewpoints, without imposing constraints to ensure that semantic features are accurately aligned with the true surfaces of objects. Consequently, the extracted semantic features are not spatially consistent with object surfaces in 3D space, which dramatically limits the application performance in downstream tasks like 3D querying, segmentation, and editing.

To address the aforementioned limitations, we propose LangSurf (Language-Embedded Surface Field). Unlike previous methods, LangSurf prioritizes the alignment of semantic features with the actual surfaces of objects in the 3D scene, ensuring a more spatially coherent semantic field. Specifically, to overcome the representation limitation of the local semantic features, our approach introduces a Hierarchical-Context Awareness Module, which first extracts pixel-level semantic features for the entire image and then applies SAM’s mask to perform mask pooling within the corresponding regions, yielding context-aware semantic feature for each mask. Such a module enriches each mask’s semantic feature by supplementing it with global context information, especially beneficial for low-texture areas or objects with intricate structures. Additionally, by retaining LangSplat's hierarchical structure, LangSurf enables the model to perceive objects at varying levels of granularity. Through this approach, LangSurf creates a more accurate and contextually aligned 3D semantic field, enabling more effective downstream applications.

LangSurf’s model architecture employs a joint learning strategy that synchronizes geometry and semantic information. We enhance the geometric quality of the semantic field through multi-view normal vector constraints, ensuring precise alignment with object surfaces. Additionally, we employ a self-supervised semantic grouping strategy that assigns language features to Gaussian points in both the 2D feature maps and 3D representations, rather than relying solely on 2D feature map supervision as in previous methods. To differentiate between objects while preserving their unique language characteristics, we implement an instance-aware training scheme that maximizes semantic distances between objects. Together, these strategies ensure precise semantic field distribution within 3D space, improving downstream task performance.
Our main contributions:

1. We introduce LangSurf, a model that emphasizes aligning 
\begin{figure*}[t]
  \centering
  \includegraphics[width=\linewidth]{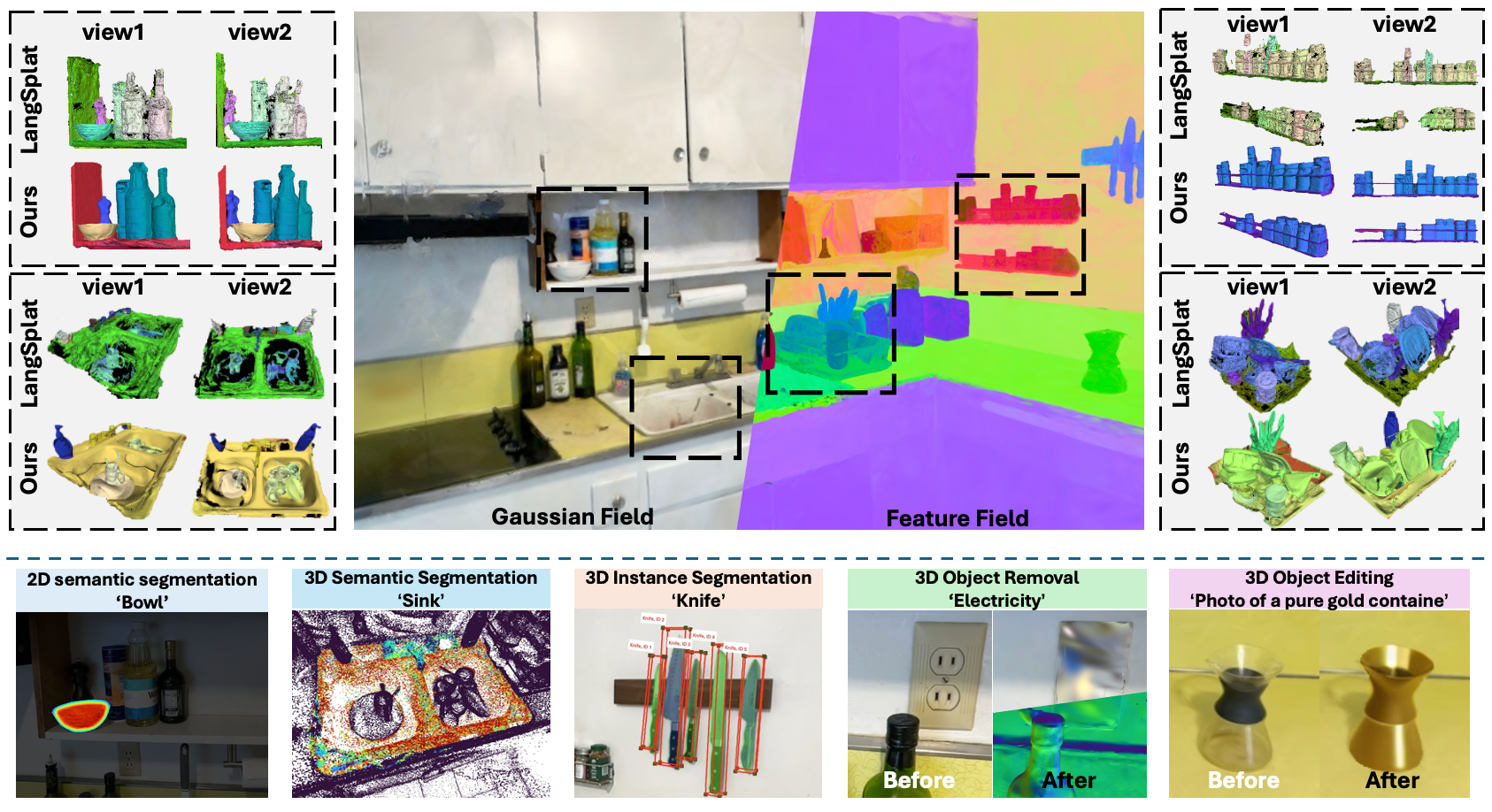}
  \caption{We proposed LangSurf, a model that aligns language features with object surfaces to enhance 3D scene understanding. The top left and right panels illustrate the qualitative differences between LangSurf and LangSplat~\cite{langsplat} in reconstructing the 3D language field from multiple viewpoints, demonstrating LangSurf's superior alignment of semantic features with object surfaces. The bottom row shows a variety of downstream applications enabled by LangSurf.}
  \label{fig:teaser}
\end{figure*}
 semantic features with the actual surfaces of objects within 3D scenes. This alignment ensures a more spatially coherent semantic field, improving the accuracy of downstream tasks such as 3D querying, segmentation, and editing.

2. We develop a Hierarchical-Context Awareness Module that extracts pixel-level semantic features from entire images and applies mask pooling using SAM's masks. This process enriches each mask's semantic feature with global context information, particularly benefiting low-texture areas and objects with intricate structures.

3. The proposed method is evaluated on the LERF and ScanNet datasets, demonstrating superior performance in open-vocabulary 2D/3D semantic segmentation tasks compared to current SOTA methods based on 2D images and 3D language fields. Additionally, the model showcases potential in 3D editing / removal applications, highlighting its versatility and effectiveness.

\section{Related Works}
    \label{sec:related}
\subsection{3D Gaussian Models}
3D Gaussian Splatting (3DGS) \cite{3dgs} has emerged as a significant advancement in 3D reconstruction, offering high-resolution real-time rendering capabilities that surpass traditional Neural Radiance Field (NeRF) methods \cite{nerf, pumarola2021d, qin2024high, park2021hypernerf, barron2021mip, barron2023zip, xu2022point}. This efficiency facilitates various downstream applications \cite{zhang2024gaussian, cai2024hdr, langsplat, wu20244d, cai2024structure, xu2024gaussian, yu2024mip}. 
In the field of 3D generation \cite{liang2024luciddreamer, cai2024baking, he2024lucidfusion}, Luciddreamer \cite{liang2024luciddreamer} incorporates 3DGS into text-to-3D generation pipeline to achieve high-quality 3D generated results. Some methods \cite{langsplat, qiu2024feature, garfield, legaussian, ji2024graspsplats, zuo2024fmgs, yue2025improving, wu2024opengaussian} focus on reconstructing 3D feature field, LangSplat \cite{langsplat} utilizes language-embedded Gaussians to enable precise and efficient open-vocabulary querying within 3D spaces.
Despite its capabilities, 3DGS struggles with accurately reconstructing 3D object surfaces. Several subsequent methods~\cite{sugar,2dgs,pgsr} enhance this process by incorporating geometric regularization. 
Sugar \cite{sugar} introduces a regularization term that encourages Gaussians to align closely with the surface of the scene. This alignment is then utilized to extract a mesh from the Gaussians through Poisson reconstruction.
PGSR \cite{pgsr} employs an unbiased depth rendering method and supervises Gaussian primitives using both single-view and multi-view loss, leading to detailed reconstruction of 3D surfaces and meshes.
Unlike these methods, our paper focuses on the surface of 3D language field, which is critical for open-vocabulary 2D / 3D segmentation, 3D removal, and editing.

\begin{figure*}[t!]
    \centering
    \includegraphics[width=0.95\linewidth]{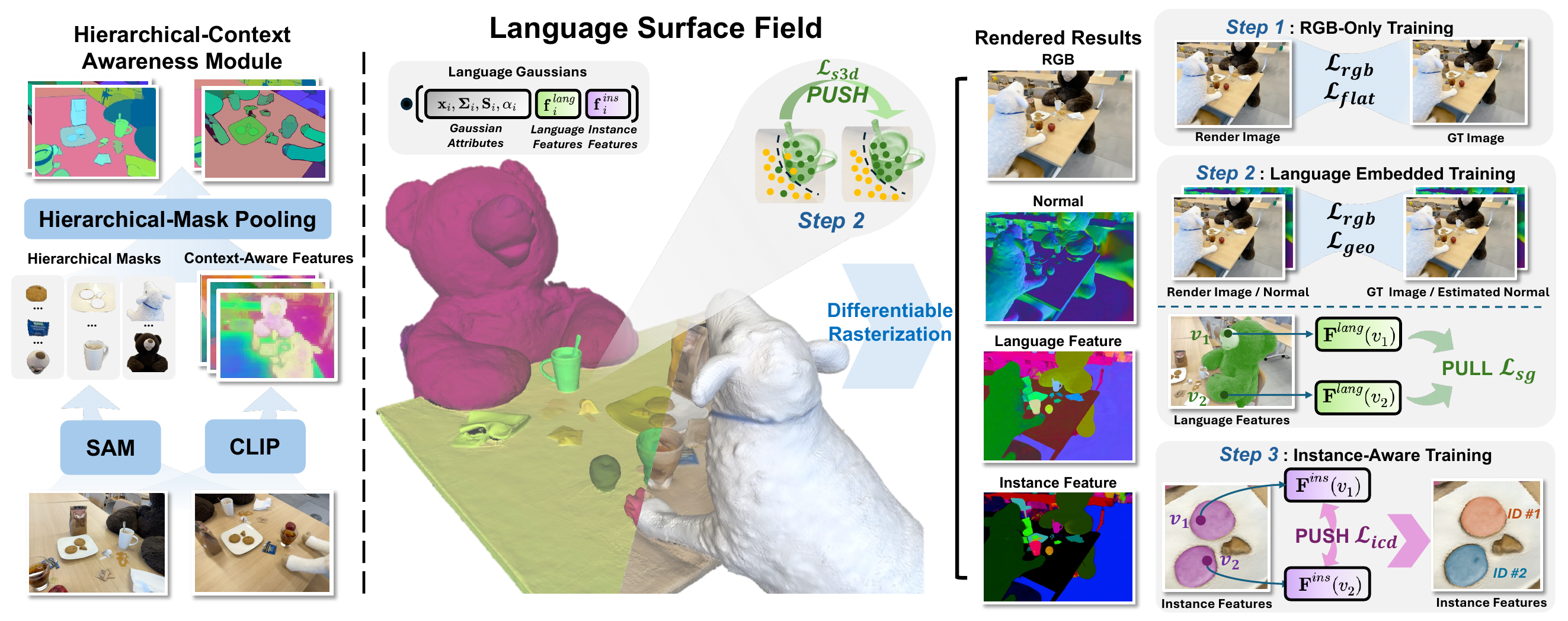}
    \caption{\textbf{Overview of proposed LangSurf.} Given input views, we reconstruct a language-embedded surface field to enable 2D / 3D open-vocabulary segmentation as well as downstream tasks. Our pipeline contains two main steps: 1) Hierarchical-Context Awareness Module extracts context-aware features with multiple hierarchies (Sec. \ref{sec:hierarchical}); 2) Language-Embedded Training utilizes a joint training strategy to construct language-embedded surface field (Sec. \ref{sec:joint}).}
    \label{fig:framework}
\end{figure*}

\subsection{3D Scene Understanding}
The integration of natural language processing with 3D scene understanding has garnered significant attention in recent years. 
Recent advancements in open-world 3D scene understanding have sought to combine Neural-based 3D representations~\cite{lerf,garfield,langsplat,gaussian-grouping,legaussian} with SAM or CLIP.
LangSplat~\cite{langsplat} advances the field by utilizing a collection of 3D Gaussians and training a feature autoencoder, each encoding language feature distilled from SAM and CLIP, to represent the scene-specific latent language field. 
Gaussian Grouping~\cite{gaussian-grouping} enhances each Gaussian with a compact identity encoding by leveraging mask predictions by SAM and spatial consistency regularization, enabling grouping based on object instance or category within the 3D scene.
However, current methods provide limited surface reconstruction, leading to inaccurate meshes. In contrast, we propose a language-embedded surface field that embeds 3D language fields with the surface to enhance surface reconstruction quality, boosting the accuracy of mesh segmentation and other geometric tasks.
\vspace{-1em}
\section{Preliminaries}
LangSplat~\cite{langsplat} integrates semantics into 3DGS for 3D scene understanding. In addition, it adds latent space \(\mathbf{f}^{lang}\in \mathbb{R}^{3}\) to Gaussian attributes to represent the language field in 3D.
Such a method facilitates real-time alpha blending of numerous Gaussians to render novel-view RGB images and language maps:
\begin{equation}
\left\{
\begin{aligned}
    \mathbf{C}(v) &= \sum_{i \in \mathcal{N}} \mathbf{c}_i \alpha_i \prod_{j=1}^{i-1}\left(1-\alpha_j\right), \\
    \mathbf{F}(v) &= \sum_{i \in \mathcal{N}} \mathbf{f}_i \alpha_i \prod_{j=1}^{i-1}\left(1-\alpha_j\right),
\end{aligned}
\right.
\end{equation}
where \(\mathbf{C}(v)\) and \(\mathbf{F}(v)\) represent the rendered color and feature at pixel \(v\), \(\mathcal{N}\) is the number of Gaussians that the ray passes through, \(\alpha\) is the opacity of the Gaussian, and \(\mathbf{c}_i \in \mathbb{R}^{3}\) is the view-dependent colors represented as a series of sphere harmonics coefficients in the practice of 3DGS. Although it can synthesize 2D feature maps from novel viewpoints, it lacks constraints to ensure that semantic features are accurately aligned with the true surfaces of objects, causing inaccurate language representation at the 3D level.

\section{Methodology}
    
Given input views  $ \{\mathbf{I}_i\in \mathbb{R}^{3\times H\times W}\}$, our main objective is to reconstruct the language-embedded surface field, which is denoted as a set of Gaussians \(\{(\mathbf{x}_i, \mathbf{\Sigma}_i, \mathbf{S}_i, \alpha_i, \mathbf{f}^{lang}_i, \mathbf{f}^{ins}_i)\}\), where   \(\mathbf{f}^{ins}\in \mathbb{R}^{3}\) denotes instance features. Such a field allows performing text queries and manipulations (\textit{i.e.} removal, editing) at the instance level. Our framework can be divided into two stages. 
\textbf{Hierarchical-Context Awareness Module} extracts language-pixel aligned features \(\{\mathbf{L}^{lang,h}_i,|h=(s,m,l)\}\) on different hierarchies from image sets to facilitate the subsequent training (see Sec. \ref{sec:hierarchical}).  
\textbf{Language-Embedded Surface Field Training} constructs the language-embedded surface field by utilizing a multi-step training strategy with 2D and 3D semantic supervision (see Sec. \ref{sec:joint}).

\begin{table*}[t]
\centering
\caption{\textbf{2D Quantitative Results on LERF Dataset.} We report the open-vocabulary localization accuracy (\%) and 2D semantic segmentation (IoU scores). LSeg~\cite{lseg} and CAT-Seg~\cite{catseg} are 2D open-vocabulary segmentation networks, while other methods~\cite{langsplat,lerf,gaussian-grouping} are language field models. We denote ``GS-Group'' as Gaussian-Grouping. The \textbf{bold} denotes the best results.}
\label{tab:lerf2dseg}
\resizebox{\textwidth}{18mm}{

\begin{tabular}{ c | c  c  | c  c  |c c | c  c  |c  c  |c  c }
\toprule
\multirow{2}{*}{Scene} & \multicolumn{2}{c|}{LSeg} & \multicolumn{2}{c|}{CAT-Seg} & \multicolumn{2}{c|}{LERF} & \multicolumn{2}{c|}{LangSplat} & \multicolumn{2}{c|}{GS-Group}  & \multicolumn{2}{c}{Ours} \\
 & mAcc\(\uparrow\) &mIou \(\uparrow\)& mAcc \(\uparrow\)&mIou \(\uparrow\)& mAcc \(\uparrow\)&mIou\(\uparrow\)& mAcc\(\uparrow\) &mIou\(\uparrow\) & mAcc\(\uparrow\) &mIou\(\uparrow\)& mAcc \(\uparrow\)&mIou \(\uparrow\)\\
\midrule
Teatime & 28.07  & 15.37  & 69.49 &34.39& 71.69 & 38.76  & \textbf{88.10} & 65.10   & 79.60 & 58.20 & 84.75 & \textbf{73.57} \\
Ramen & 8.45 & 2.16  & 53.52 &16.77 & 54.71 & 21.54 & 56.34 & 46.52   & 30.90 & 24.30 & \textbf{63.40} & \textbf{47.03} \\
Kitchen & 44.00 & 20.87  & 68.00 &28.24 & 64.15 & 29.19 & 72.73 & 50.83   & 54.50 & 39.40 &\textbf{81.82} & \textbf{54.99} \\
Bouquet & 47.83 & 14.87 & 78.26& \textbf{44.31} & 72.28 & 35.69 & 43.48 & 23.29 & 90.09 & 00.95 & \textbf{91.30} & 40.45 \\
Hand\_Hand & 35.00 & 27.85 & 88.00 & 58.71 & 86.42 & 37.89 & 85.00 & 72.12 & 36.20 & 26.77 & \textbf{90.00} & \textbf{75.69}\\
Donuts & 65.38 & 31.80 & 95.00 & 46.17 & 87.13 & 39.91 & \textbf{97.00} & 53.57 & 25.20 & 23.04 & 96.15 & \textbf{67.50} \\ \midrule
Overall & 38.12 & 18.82 & 75.38 & 38.10 & 63.51 & 29.83 & 74.28 & 51.90 & 52.76 & 28.78 & \textbf{84.57} & \textbf{60.02} \\
\bottomrule

\end{tabular}
}
\end{table*}

\subsection{Hierarchical-Context Awareness Module}
\label{sec:hierarchical}
In order to correctly capture visual-language aligned features, we propose a simple but efficient method named Hierarchical-Context Awareness Module. 
Unlike previous work~\cite{langsplat} extracted language features for the masked objects segmented by SAM, for each input image \(\mathbf{I}_i\), we apply a pre-trained image encoder~\cite{openseg} to produce image-wise features \(\mathbf{L}_i^{lang}\in\mathbb{R}^{D\times H \times W}\). It considers contextual information to enable accurate visual-language alignment of the features, especially for obscured objects and objects with low textures and shapes. 

However, simply adopting \(\mathbf{L}_i^{lang}\) to train the Gaussian model will limit the ability of open-vocabulary query with different scales since context-aware features suppress the diversity of the features, making it difficult to distinguish small objects from the large one, such as ``bear nose'' and ``bear''.
To this end, we perform Hierarchical-Mask Pooling for the image features, which uses the multi-hierarchy masks \(\{\mathbb{M}_i^h, | h = s,m,l\}\) segmented by SAM to decompose the multi-scale language features from the original features \(\mathbf{I}_i\), where \(s,m,l\) represents the small, medium, and large hierarchy and each hierarchy obtains multi masks \(\mathbb{M}_i^h = \{\mathbf{M}_{i,j}^h, (j=1,\cdots,M)\}\).

\begin{table}[t]
\centering
\caption{\textbf{2D Quantitative Results on ScanNet Dataset.} We report the open-vocabulary 2D semantic segmentation (IoU scores). The \textbf{bold} denotes the best results.}
\resizebox{0.48\textwidth}{14mm}{
\begin{tabular}{ c  |c  |c  |c  |c  |c }
\toprule
Scene& LSeg & CATSeg & LangSplat & GS-Group & Ours \\
\midrule
085 & 32.75 & 42.36 & 36.69 & 43.41 & \textbf{66.89} \\

114 & 19.11 & \textbf{33.66} & 16.81 & 20.78 & 33.25 \\

616 & 33.83 & 45.57 & 24.29 & 23.84 & \textbf{51.40} \\

617 & 15.53 & 24.51 & 11.49 & 19.42 & \textbf{40.09} \\
\midrule
Mean & 25.30 & 36.52 & 22.32 & 26.86 &  \textbf{47.91} \\
\bottomrule
\end{tabular}
}
\label{tab:scannet2dseg}
  \vspace{-0.15in}
\end{table}

For each hierarchy, image features perform masked average pooling to enhance semantic consistency within the masks:
\begin{equation}
    \mathbf{L}^{lang, h}(v) = \frac{\sum \mathbf{L}^{lang}(v) \cdot  \mathbf{M}^h(v)}{\sum \mathbf{M}^h(v)}, h = \{s,m,l\},
\end{equation}
where \(v\) represents the pixel within the mask region at the hierarchy \(h\).
%
%

In the end, we adopt an end-to-end autoencoder that compresses the language features $\{\mathbf{H}^{lang,h}\}$ into low-dimensional latent space \(\{\mathbf{L}^{lang,h}\}\) during training and decodes the latent space into original features during inference, where \(\{\mathbf{L}^{lang,h}_i\}\) is a 3-channels feature map. It reduces memory consumption and improves efficiency. 

\subsection{Language-Embedded Surface Field Training}
\label{sec:joint}
%
%
We decompose the training procedure into three steps: 1) basic RGB supervision is deployed to obtain basic 3D representation; 2) both geometry and semantic supervision are deployed to optimize the language Gaussians with not only accurate spatial semantic distributions; 3) well-trained language features initialize the instance features of Gaussians, then instance-level training are deployed to distinguish objects from the language space.
These stages are as follows:

\begin{table}[t]
\centering
\caption{\textbf{3D Quantitative Results on ScanNet Dataset.} We report the average open-vocabulary Semantic F-Score\(\uparrow\). The \textbf{bold} denotes the best results.}
\label{tab:scannet3dseg}
\resizebox{0.45\textwidth}{14mm}{
\setlength{\tabcolsep}{4.5mm}{
\begin{tabular}{c | c | c | c}
\toprule
Scene & LangSplat & GS-Group & Ours \\
\midrule
085 & 6.19 & 18.92 & \textbf{52.70} \\
114 & 13.19 & 13.35 & \textbf{30.62} \\
616 & 08.12 & 09.47 & \textbf{36.44} \\
617 & 11.39 & 10.62 & \textbf{33.05} \\
\midrule
Mean & 9.72 & 13.09 & \textbf{38.20} \\
\bottomrule
\end{tabular}
}}
\vspace{-3mm}
\end{table}

\noindent \textbf{Step 1: RGB-Only Training.} In this step, we aim to obtain the basic 3D field by deploying basic RGB supervision \(\mathcal{L}_{rgb}\) followed by a Gaussian flatten supervision \(\mathcal{L}_{flat}\):
\begin{equation}
\left\{
\begin{alignedat}{2}
    &\mathcal{L}_{rgb}  &&= \|\mathbf{C}_i - \mathbf{I}_i \|_1, \\
    &\mathcal{L}_{flat} &&= \left\|\min \left(s_1, s_2, s_3\right)\right\|_1,
\end{alignedat}
\right.
\end{equation}
where \(\{s_1, s_2, s_3\} = diag(\mathbf{S}_i)\) are the scale factors of Gaussian. It compresses the Gaussians and makes them flatten into the object planes.

\noindent \textbf{Step 2: Language-Embedded Training.} Here, we perform joint training using geometry and semantic supervision to construct a language-embedded surface field. Firstly, following the strategy of PGSR~\cite{pgsr}, we adopt geometry regularization constraints \(\mathcal{L}_{geo}\) to optimize the geometry representation of our model.
Apart from regular L2 semantic loss between $\mathbf{F}^{lang}$ and $\mathbf{H}^{lang}$, we further propose Semantic Grouping terms \(\mathcal{L}_{sg}\), which groups the rendered features \(\mathbf{F}^{lang}(\cdot)\) within the same mask \(\mathbf{M}_j\) of the image by minimizing their semantic distance. Such a strategy maintains semantic consistency within the object and creates clearer boundaries between different objects, the formulation is shown below:
\begin{equation}
    \mathcal{L}_{sg}=\frac{1}{M }\sum_{j=1}^M \sum_{v_1, v_2 \in \mathbf{M}_j}\left\|\mathbf{F}^{lang}(v_1)-\mathbf{F}^{lang}(v_2)\right\|_2,
\end{equation}
where \(M\) is the total numbers of segmentation masks \(\mathbb{M} = \{\mathbf{M}_{j}, (j=1,\cdots,M)\}\) of the image. 
Meanwhile, given the fact that the outlier language Gaussians seriously disturb the 3D localization of the objects as well as downstream tasks, we additionally propose a Spatial-Aware Semantic Supervision \(\mathcal{L}_{s3d}\) to increase the spatial constraint of language Gaussians to suppress the outlier of language Gaussians. 
It utilizes KL-divergence supervision to align the semantic features with the top-k nearest Gaussians:
\vspace{-0.1in}
\begin{equation}
    \mathcal{L}_{s3d} = \sum_{j=1}^{N} \sum_{k=1}^{N_k}\mathbf{f}^{lang}_j \cdot\left(\log \left(\mathbf{f}^{lang}_j / \mathbf{f}^{lang}_k\right)\right),
\end{equation}
where \(N\) denotes the total number of Gaussian points and \(N_k\) is the Top-K nearest number of the \(i\)-th Gaussian point.
Such strategies enable our language-embedded surface field to fit into the surface of the scene accurately, achieving comprehensive scene understanding.

\begin{figure}[t]
  \centering
  \includegraphics[width=\linewidth]{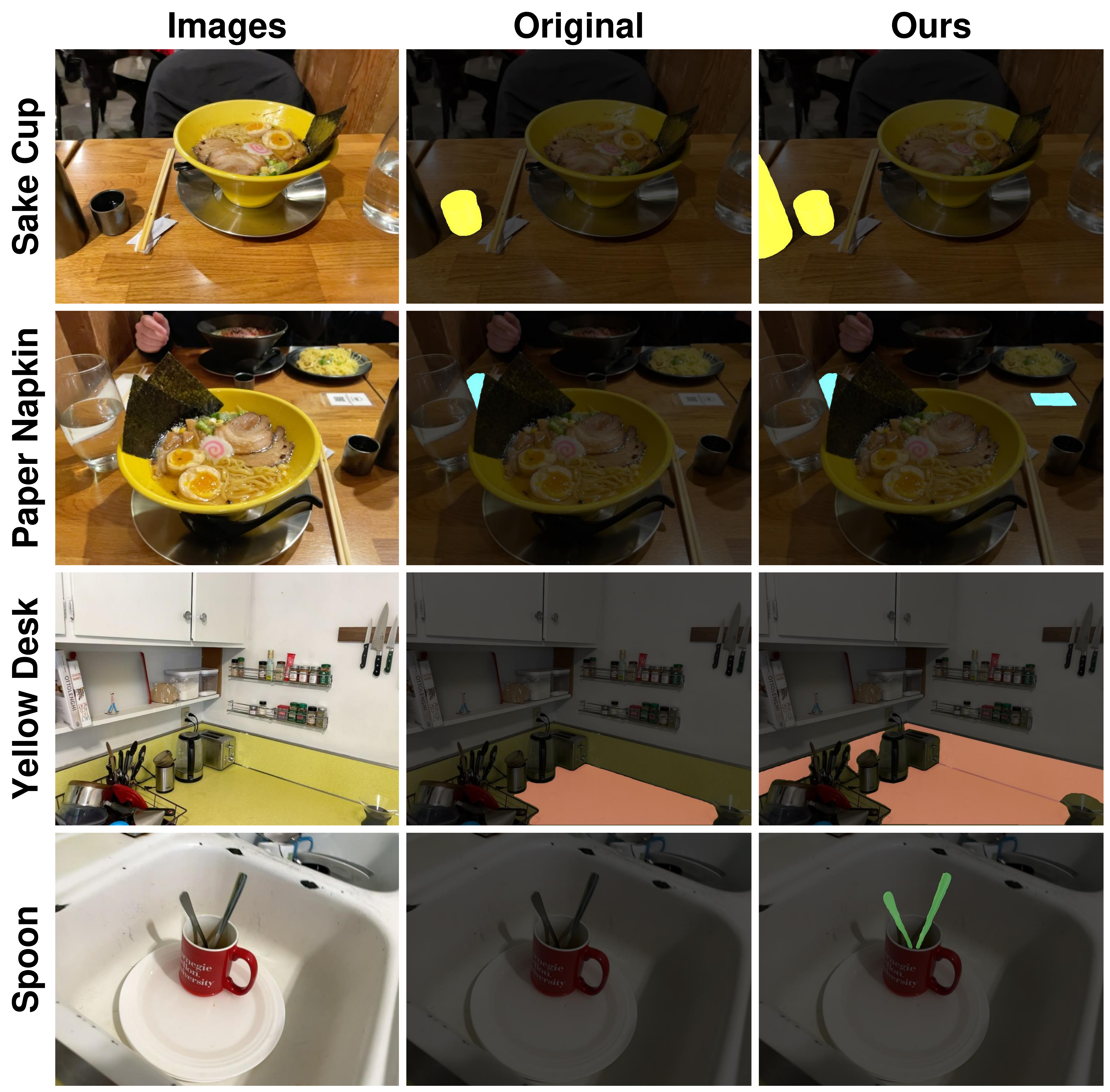}
  \caption{\textbf{Visualization of our refined annotations in LERF dataset~\cite{lerf}}. We showcase some differences between the original annotations and our refined annotations.}
  \label{fig:supp_dataset}
\end{figure}

\begin{figure}[t]
  \centering
  \includegraphics[width=\linewidth]{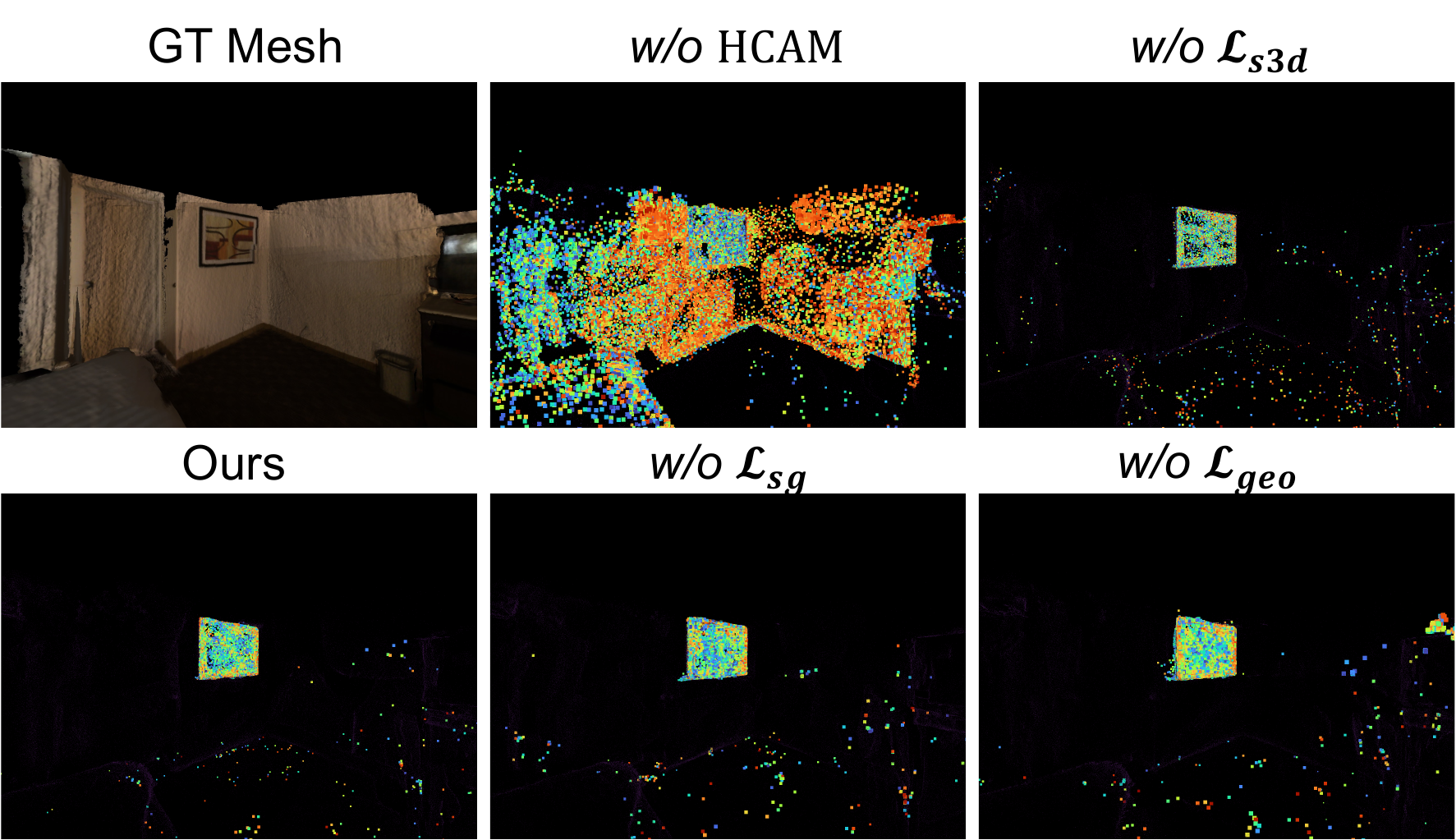}
  \vspace{-0.2in}
  \caption{\textbf{Qualitative Results of Ablations on  ScanNet~\cite{scannet} Dataset.} We visualize the 3D open-vocabulary segmentation with ``pictures'' prompt.}
  \label{fig:ablation}
  \vspace{-0.3in}
\end{figure}

\begin{figure}[t]
  \centering
  \includegraphics[width=\linewidth]{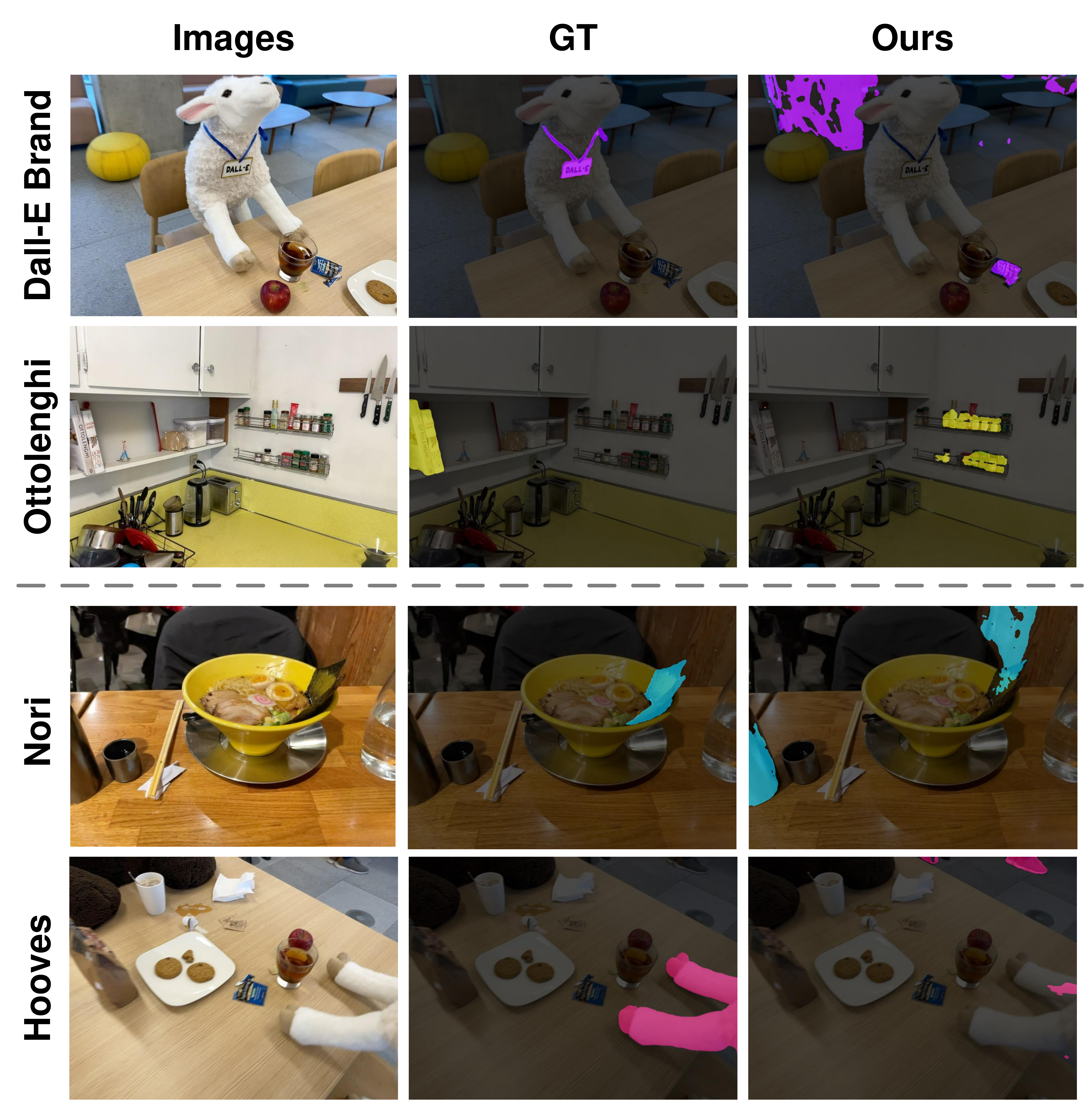}
  \caption{\textbf{Visualization of failure cases in LERF dataset~\cite{lerf}}. Top: failure cases where texts in the images cannot be detected. Bottom: failure cases where objects of the tail class cannot be recognized.}  \label{fig:supp_failure}
  \vspace{-0.1in}
\end{figure}

\noindent \textbf{Step 3: Instance-Aware Training.} 
We expand our language-embedded surface field for perceiving instance level since there are multiple objects with the same categories in the scenes.
To this end, we introduce new instance-aware parameters to assign each Gaussian to its corresponding instance or stuff in the 3D scene while preserving all attributes of the language Gaussians for text-aligned querying.
In detail, we use well-trained language features  \(\mathbf{f}^{lang}_i\) to initialize the instance features \(\mathbf{f}^{ins}_i\) for each Gaussian.
After that, we compute the mean feature  \(\mathbf{z}^{ins}_i\) for each masked region \(\mathbf{M}_i\) on the render instance map \(\mathbf{F}^{ins}_i\):
\begin{equation}
    \mathbf{z}^{ins}_i=\frac{1}{\left|\mathbf{M}_i\right|} \sum_{v \in \mathbf{M}_i} \mathbf{F}^{ins}_i(v),
\end{equation}
where \(\left|\cdot\right|\) represents the area of the mask. In the end, an Instance Contrastive Decomposition supervision \(\mathcal{L}_{icd}\) is proposed to decompose the objects by maximizing the distance between instance features of different masks:
\begin{equation}
    \mathcal{L}_{icd}=\sum_{j=1}^M \sum_{k\neq j}^M\text{ReLU}(D_{min} -\left\|\mathbf{z}^{ins}_j-\mathbf{z}^{ins}_k\right\|_2),
\end{equation}
where \(D_{min}\) is the minimum distance between instances.  Notably, we only train the instance features of the Gaussians instead of all parameters.
\\Instance-aware features enable two key downstream applications:object removal task and object editing task. For the first task, we reconstruct the scene with our language-embedded surface field, perform a text-guided query for language Gaussians and select the Gaussians with high activation scores.  With the selected Gaussians, we compute the convex hull of the selected Gaussians and then identify all Gaussians from the original Gaussians that are inside this convex hull.   After that, we delete the identified Gaussians.For the second task, we reconstruct the scene with our language-embedded surface field and perform a text-guided query for language Gaussians and select the Gaussians with high activation scores.After that, we render novel view object masks, RGB images, and depth maps of the edited object, and then feed them into the GaussCtrl~\cite{gaussctrl2024} to generate edited images.In the end, we finetune our Gaussian model for 3,000 steps using the edited images.

\section{Experiments}

\subsection{Implementation Details}
\subsubsection{Training Details}
To extract image features of each image, we utilize the backbone of OpenSeg~\cite{openseg} as the extractor with a Vallina CLIP for the text encoder. For SAM~\cite{sam}, we use the ViT-H model to segment 2D masks. During Language-Embedded Surface Field training, we train 7,000 iterations for Step 1; 23,000 iterations for Step 2; 10,000 for Step 3.
To evaluate the comprehensive scene understanding of our proposed language-embedded surface field, we employ two datasets, LERF~\cite{lerf} and ScanNet~\cite{scannet}. 
The LERF dataset is an in-the-wild dataset captured by a handheld device, which consists of a greater number of object elements. 
It also consists of precise annotations on both segmentation masks and bounding boxes at 2D level. 
The ScanNet dataset is a large dataset captured by RGB-D devices in complex indoor scenes. 
Each scene includes precise poses and semantic labels at 3D level. These characteristics make it suitable for 3D scene understanding tasks. 

\subsubsection{Evaluation Metrics}

\begin{figure*}[t]
  \centering
  \includegraphics[width=1\linewidth]{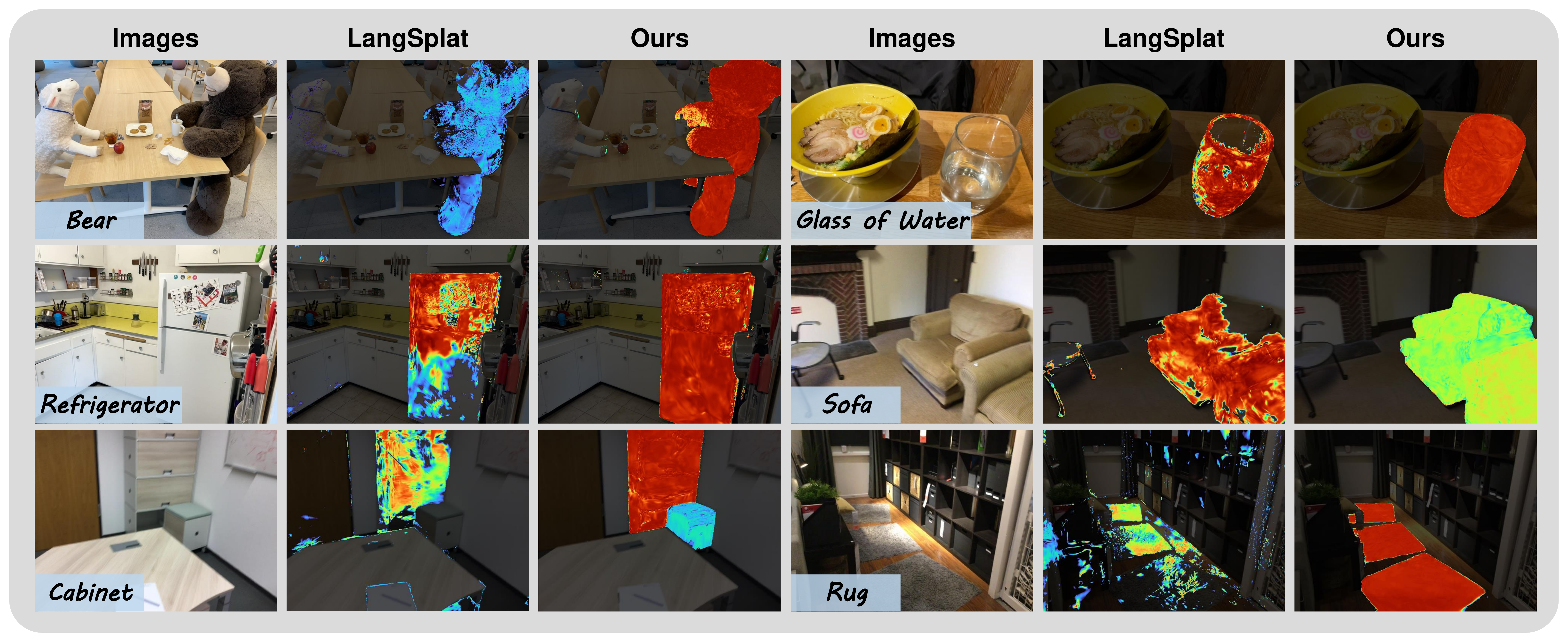}
  \caption{\textbf{Visualization of our 2D text-query heatmaps in both datasets.}. We showcase our results compared with LangSplat~\cite{langsplat}.}
  \label{fig:supp_heatmap}
\end{figure*}

\begin{figure*}[t]
  \centering
  \includegraphics[width=1\linewidth]{figures/pcmesh.pdf}
  \vspace{-0.2in}
  \caption{\textbf{3D Qualitative Results on both LERF~\cite{lerf} and ScanNet~\cite{scannet} Datasets.} We compare our method with other GS-based methods~\cite{langsplat,gaussian-grouping}. We show the queried point clouds with activated score and mesh corresponding to the given text.}
  \label{fig:pc_mesh}
\end{figure*}

\begin{table*}[t]
\centering
\caption{\textbf{Quantitative Results of ScanNet~\cite{scannet}.} We present Semantic F-Score for each text.}
\footnotesize
\begin{tabular}{c|cccccccccccccc}
\toprule
\multirow{2}{*}{Methods} & \multicolumn{7}{c|}{scene0085\_00}                              & \multicolumn{7}{c}{scene0616\_00}                              \\
                         & bed    & chair  & curtain & picture & wall   & floor  & \multicolumn{1}{c|}{door}   & wall   & floor  & chair  & table   & door   & window & picture \\ \midrule
LangSplat                & 08.68 &   00.00   & 15.8   & 02.97  & 13.44 & 01.83 & \multicolumn{1}{c|}{00.61} & 10.34 & 03.98 & 28.79 & 04.55  & 01.02 & 02.91 & 05.26  \\
GS-Group                 & 45.24 & 06.62 & 24.32  & 05.19  & 15.96 & 34.99 & \multicolumn{1}{c|}{00.18} & 04.61 & 05.56 & 26.03 & 14.6   & 04.16 & 01.32 & \textbf{10.01}  \\
Ours                     & \textbf{65.13} & \textbf{87.05} & \textbf{60.39}  & \textbf{53.82}  & \textbf{29.72} & \textbf{55.17} & \multicolumn{1}{c|}{\textbf{17.63}} & \textbf{12.98} & \textbf{45.09} & \textbf{74.43} & \textbf{66.72}  & \textbf{23.03} & \textbf{27.76} & 06.32  \\ 
\end{tabular}
\begin{tabular}{c|cccccccc|cccccc}
\toprule
\multirow{2}{*}{Methods} & \multicolumn{8}{c|}{scene0114\_02}                                       & \multicolumn{6}{c}{scene0617\_00}                     \\
                         & wall   & floor  & cabinet & chair   & table  & door   & window & \multicolumn{1}{c|}{desk}   & wall   & floor  & cabinet & sofa   & table  & curtain \\ \midrule
LangSplat                & 05.26 & 01.11 & 22.15  & 37.48  &   00.00   & 01.33 & 02.06 & \multicolumn{1}{c|}{36.13} & 00.49 & 09.54 & 04.18  & 42.01 & 11.56 & 00.57  \\
GS-Group                 & 06.01 & 02.81 &     00.00  & 38.21  & 00.83 & 01.91 & 08.97 & \multicolumn{1}{c|}{34.72} & 00.4  & 15.38 &  00.00     & 30.26 & 17.07 & 00.64  \\
Ours                     & \textbf{15.19} & \textbf{19.11} & \textbf{48.17}  & \textbf{74.93}  & \textbf{08.67} & \textbf{01.98} & \textbf{21.55} & \textbf{55.84} & \textbf{08.52} & \textbf{50.81} & \textbf{23.68}  & \textbf{72.02} & \textbf{26.97} & \textbf{17.58}  \\ \bottomrule
\end{tabular}
\label{tab:supp}
\end{table*}

\noindent\textbf{2D evaluation.} For LERF~\cite{lerf} dataset, we adopt mIoU and localization mAcc for open-vocabulary semantic segmentation. For localization mAcc, we begin by applying a mean convolution filter of size 20 to smooth the values in the relevancy maps, which helps reduce the impact of outliers. We then identify the maximum score within the relevancy maps and consider it accurately localized if its coordinates fall within the ground truth bounding box. In the end, we calculate the average accuracy across all classes.

\noindent\textbf{3D evaluation.} For 3D open-vocabulary segmentation on ScanNet, we adopt Semantic F-Score for evaluation. We select each text $t \in T$ to query all points. For the queried points $P_t$ and ground truth points $P_t^*$, the Semantic F-Score is computed as follows:
\begin{equation}
\begin{aligned}
    Precision_t &= \text{mean}_{p \in P_t}\left(\min_{p^* \in P_t^*}\Vert p - p^* \Vert < \tau \right), \\
    Recall_t &= \text{mean}_{p^* \in P_t^*}\left(\min_{p \in P_t}\Vert p - p^* \Vert < \tau \right), \\
    S\text{-}F\text{-}Score_t &= \frac{2 \times Precision_t \times Recall_t}{Precision_t + Recall_t}.
\end{aligned}
\end{equation}
where $S~F\mbox{-}Score_t$ is Semantic F-Score for text $t$ and $\tau$ is a threshold which we set to 0.05 here.

\subsubsection{Dataset Refinement}
In the original LERF dataset, multiple objects with the same categories are ignored in the previous annotations, which leads to inaccurate evaluations.
To this end, we refine the annotation errors in LERF datasets for comprehensive evaluation, as shown in Fig.~\ref{fig:supp_dataset}.
Notably, we don't introduce new categories or new images during the refining procedure.

\begin{table*}[t]
\setlength{\tabcolsep}{10pt}
\centering
\caption{\textbf{Ablation of Proposed Components in ScanNet.} We present the Semantic F-Score for each text in the 3D open-vocabulary task.}
\begin{tabular}{c|ccccccc|c}
\toprule
 Components & bed& chair& curtain& picture& wall& floor&door &Overall\\ \midrule
 \textit{w/o} HCAM                 & 57.65 & 73.25 & 34.16 & 05.27 & 24.87 & 17.04 & 01.62 & 30.55  \\
 \textit{w/o} \(\mathcal{L}_{geo}\)& 59.94 & 92.39 & 59.82 & 50.75 & 27.91 & 45.21 & 11.13 & 49.59 \\
 \textit{w/o} \(\mathcal{L}_{sg}\) & 62.18 & 86.59 & 62.95 & 55.58 & 25.95 & 52.45 & 10.54 & 50.89 \\
 \textit{w/o} \(\mathcal{L}_{s3d}\)& 64.85 & 84.28 & 65.52 & 50.47 & 26.00 & 51.60 & 08.18 & 50.13 \\
 \textit{w/}  all            & 64.80 & 88.97 & 59.04 & 61.91 & 25.14 & 51.70 & 11.51 & \textbf{51.87} \\ \bottomrule
\end{tabular}
\label{tab:supp_abla}
\end{table*}

We also demonstrate the details of our proposed components in ScanNet~\cite{scannet}. In the 3D open-vocabulary task, we provide the Semantic F-Score for each text in Tab.~\ref{tab:supp_abla}. We observe that the Hierarchical-Context Awareness Module (HCAM) makes significant improvements. Additionally, $\mathcal{L}_{geo}$, $\mathcal{L}_{sg}$ and $\mathcal{L}_{s3d}$ are complementary, with performance drops when removing either one.

\begin{figure}[t]
  \centering  \includegraphics[width=1\linewidth,height=0.25\textheight]{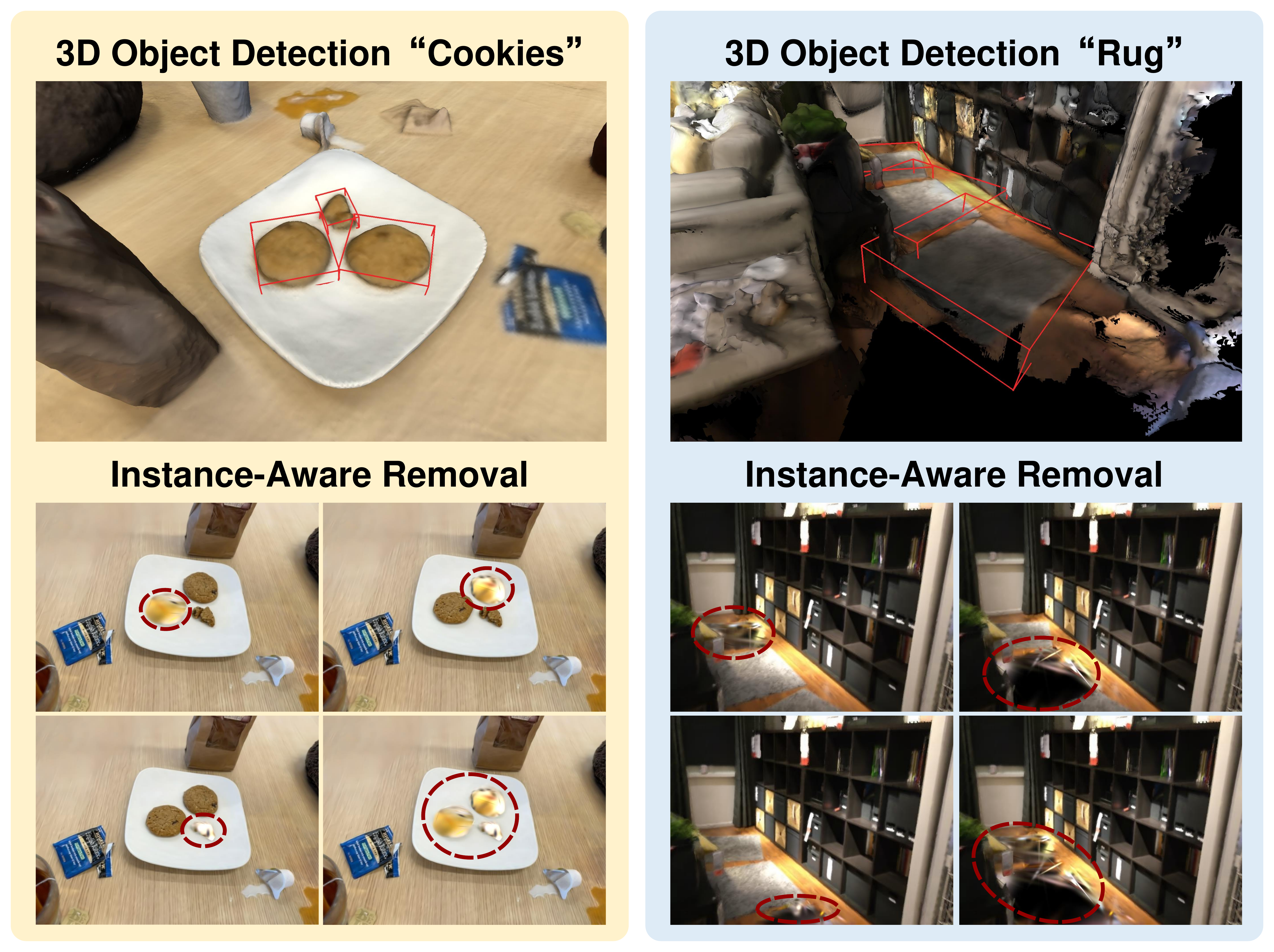}
  \vspace{-0.2in}
  \caption{Qualitative Results of \textbf{Instance-Level 3D Object Detection and Removal}. Objects are highlighted with '{\color[HTML]{CB0000}$\bigcirc$}'.}
  \label{fig:remove_instance}
\end{figure}


\begin{figure}[t]
  \centering
  \includegraphics[width=1\linewidth]{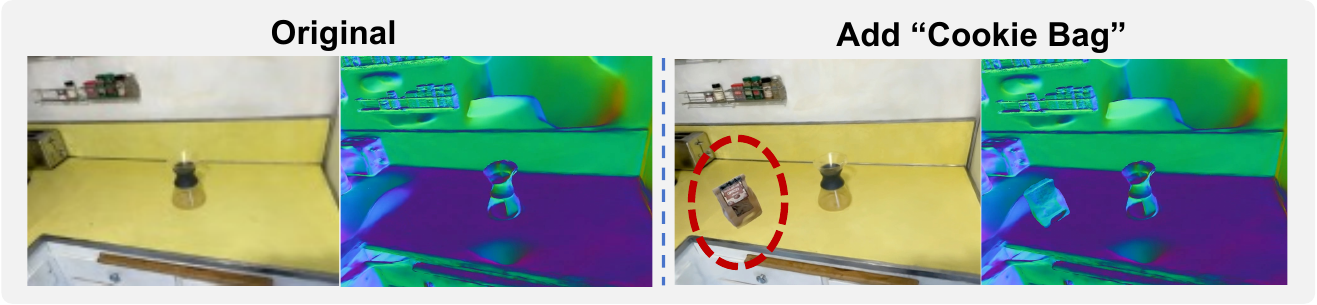}
  \vspace{-0.2in}
  \caption{\textbf{Qualitative Results of 3D Object Adding on LERF Datasets.} Objects are highlighted with '{\color[HTML]{CB0000}$\bigcirc$}'.}
  \vspace{-0.2in}
  \label{fig:add}
\end{figure}


\subsection{Applications}
We apply our method to two representative downstream tasks (\textit{i.e.} Object Removal and Editing) to validate the performance of our language-embedded surface field.

\noindent \textbf{3D Scene Objects Removal \& Adding.}
We employ text-query-based 3D object removal tasks at both semantic and instance levels.
As shown in Fig.~\ref{fig:remove} and Fig.~\ref{fig:remove_instance}, our method effectively removes the text-queried items without affecting nearby scenes compared with Gaussian Grouping. Meanwhile, Fig.~\ref{fig:add} demonstrates adding queried objects (\textit{i.e.} cookie bag) from "Teatime" scene to "Kitchen" scene, which reflects our method achieves superior language representation in 3D.

\noindent \textbf{3D Scene Objects Editing.}
Object Editing aims to edit the Gaussians of the queried object in 3D scene.
Fig.~\ref{fig:editing} shows our method enables precisely editing the Gaussians of the object while minimizing any impact on the background, which reflects our language-embedded surface field has accurate language 3D representation.

\vspace{-0.1in}


\section{Conclusions}
In this paper, we propose Language-Embedded Surface Field that accurately represents the language field in 3D to align the surface of objects. 
Unlike previous methods that focus on 2D render tasks, we propose a joint training strategy that uses geometry constraint and semantic contrastive losses to align the language field with the surface of objects to enhance the 3D representation of our model.
Furthermore, a Hierarchical-Context Awareness Module is proposed to extract context-aware features in different hierarchies for the Gaussian model training.
Comprehensive experiments demonstrate that LangSurf achieves significant performance improvements on both 2D and 3D metrics (sometimes $>$ 10\%) compared with existing SOTA methods.  Based on this, LangSurf can further support multiple downstream tasks, including object removal and editing, which also achieves significant improvements compared with existing methods.

\begin{figure*}[t]
  \centering  \includegraphics[width=0.8\linewidth]{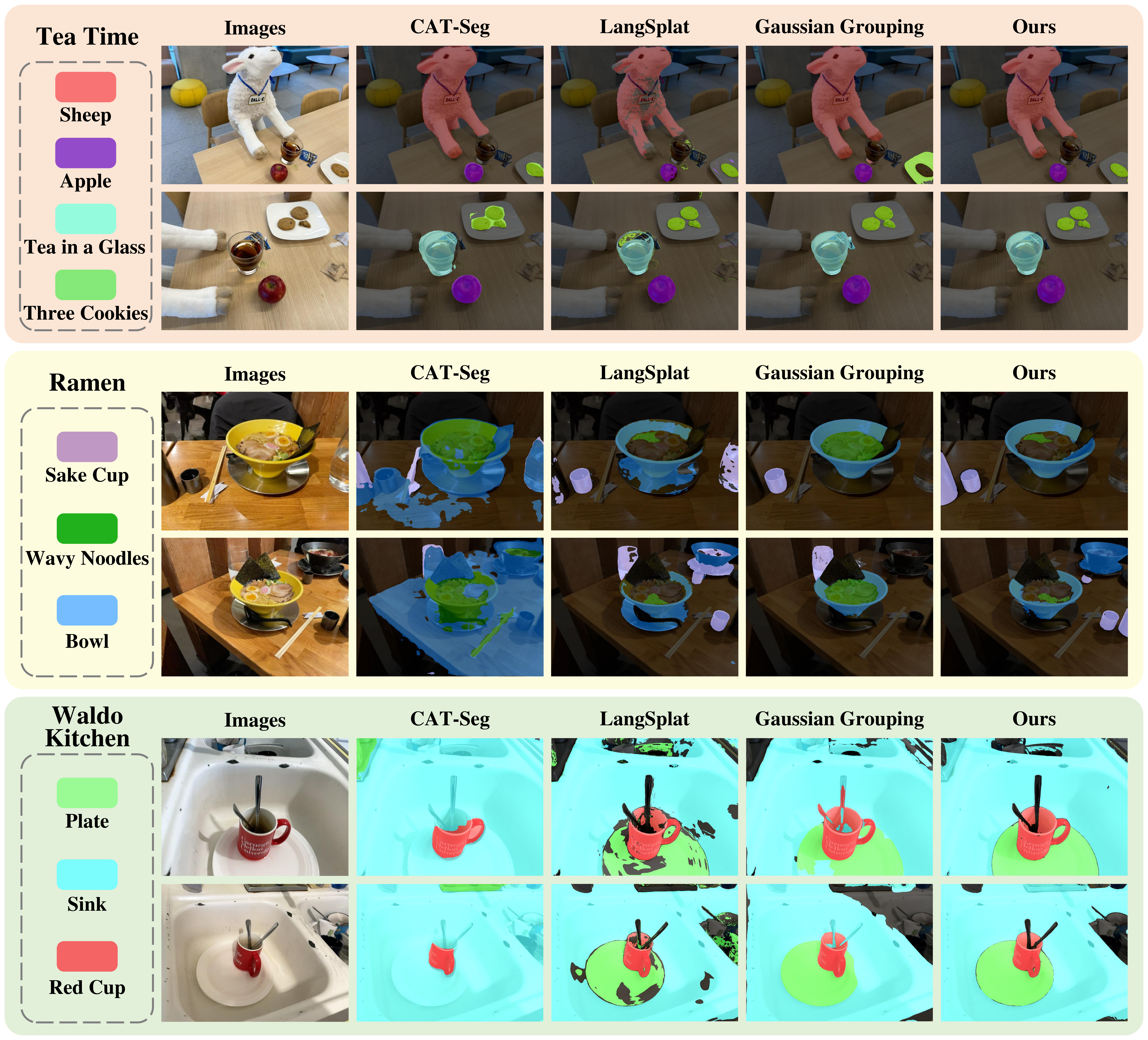}
  \caption{\textbf{Visualization of our 2D open-vocabulary masks in LERF dataset~\cite{lerf}}. Here we showcase three scenes (\textit{i.e. }Teatime, Ramen, Waldo Kitchen) with multiple text-query segmentation masks. On the left, we present the images alongside the queried texts. On the right, we display the rendered results of our method and other methods.}
  \label{fig:supp_lerf}
\end{figure*}

\bibliographystyle{IEEEtran}
\bibliography{example_paper}

\newpage
 

\begin{IEEEbiography}[{\includegraphics[width=1in,height=1.25in,clip,keepaspectratio]{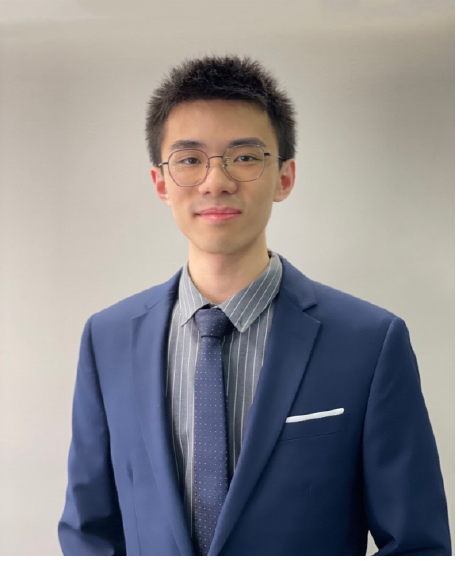}}]{Hao Li}
received the B.E. degree from Beijing University of Chemical Technology, Beijing, China, in 2022. He is currently working toward the Ph.D. degree in the School of Automation, Northwestern Polytechnical University, Xi’an, China. His research interests include computer vision and pattern recognition, especially on weakly supervised object segmentation, instance segmentation and 3D scene understanding.
\end{IEEEbiography}

\begin{IEEEbiography}[{\includegraphics[width=1in,height=1.25in,clip,keepaspectratio]{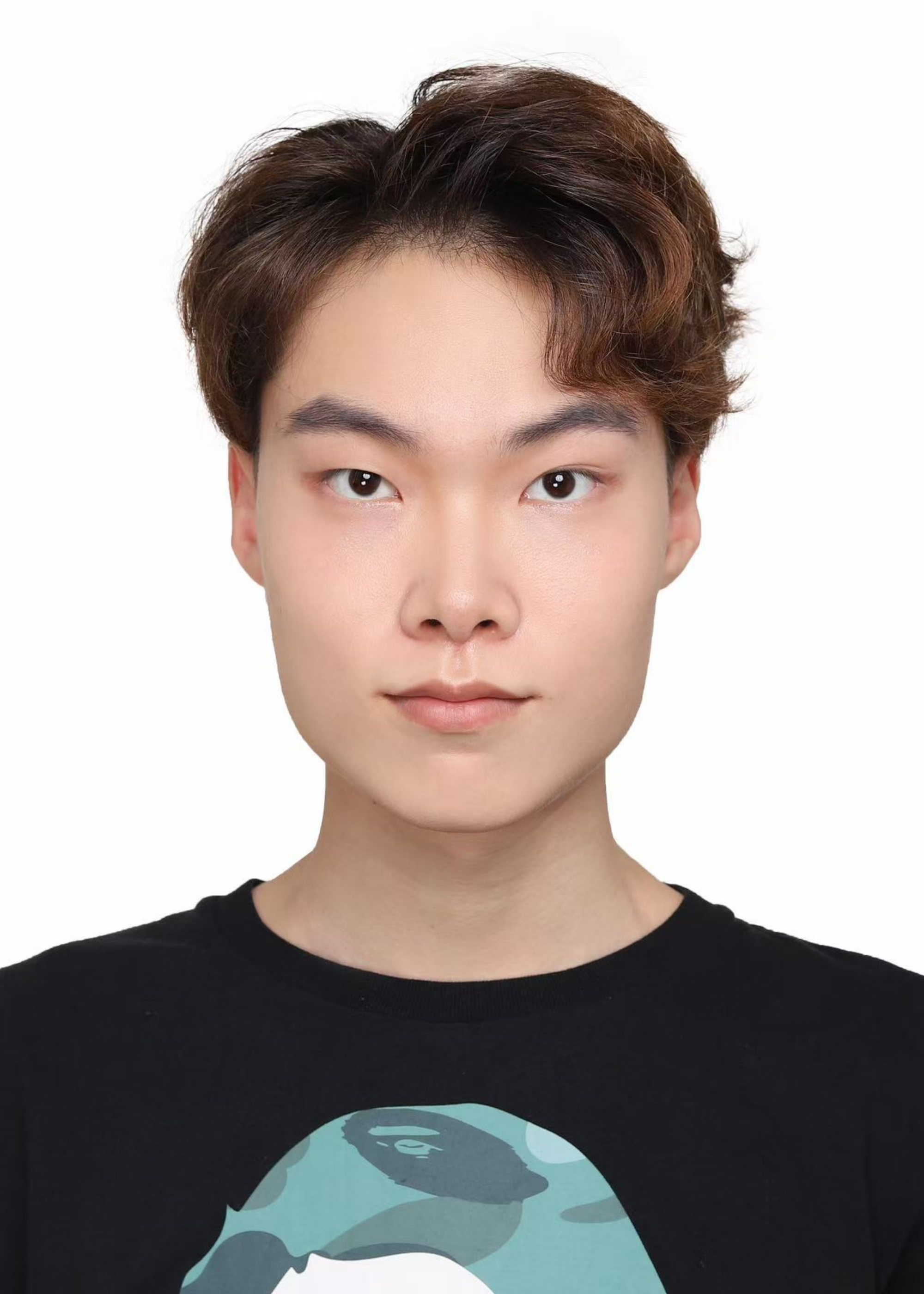}}]{Minghan Qin}
received the B.E. degree from Southeast University, China, in 2021. He received the M.E. degree in artificial intelligence from Tsinghua University, China, in 2024. His research interests include 3D perception, robotics, VLM and world model.
\end{IEEEbiography}

\begin{IEEEbiography}[{\includegraphics[width=1in,height=1.25in,
clip,keepaspectratio]{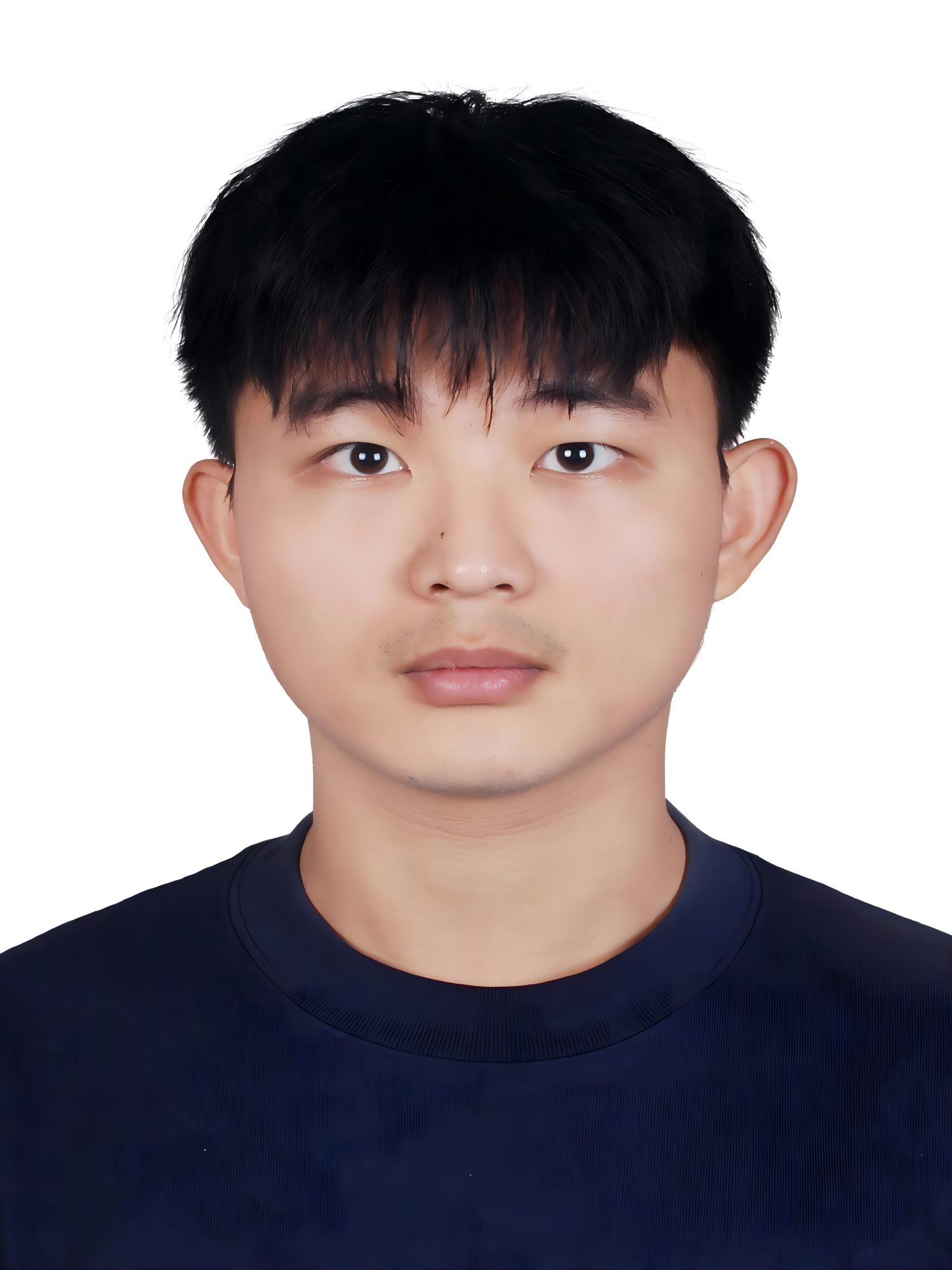}}]{Zhengyu Zou}
 received the B.E. degree from Hunan University, Changsha, China, in 2024. He is currently working toward the M.E. degree in the School of Automation, Northwestern Polytechnical University, Xi'an, China. His research interests include computer vision, novel view synthesis, 3D scene representation and reconstruction, and vision-based localization.

\end{IEEEbiography}

\begin{IEEEbiography}[{\includegraphics[width=1in,height=1.25in,clip,keepaspectratio]{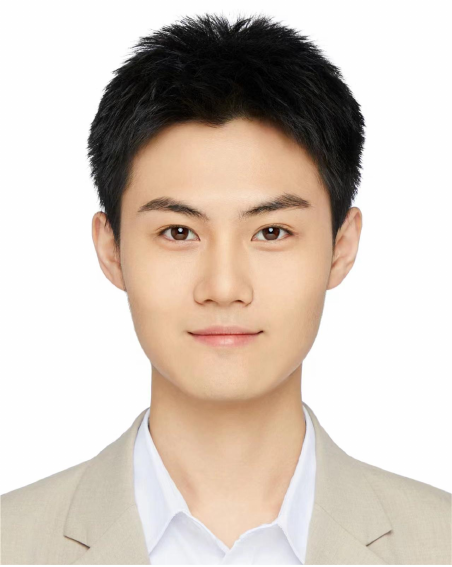}}]{Diqi He}
received the B.E. degree from Northwestern Polytechnical University, Xi'an, China. He is currently working toward an M.E. degree in the School of Automation, Northwestern Polytechnical University, Xi’an, China. His research interests include computer vision and pattern recognition, especially on vision segmentation and large scale models.
\end{IEEEbiography}

\begin{IEEEbiography}[{\includegraphics[width=1in,height=1.25in,clip,keepaspectratio]{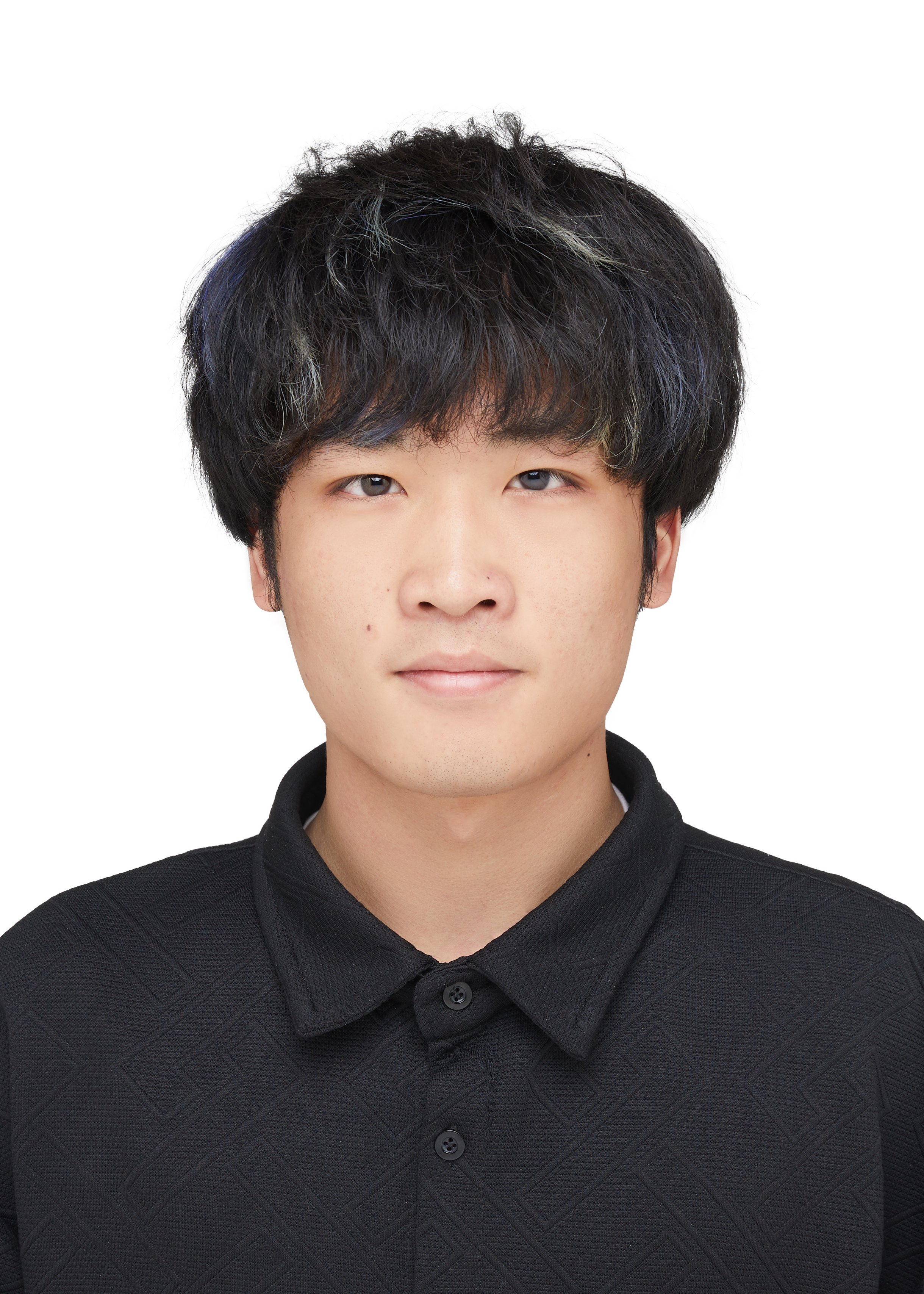}}]{Xinhao Ji}
is currently a undergraduate student at Peking University, Beijing, China, and an intern at Northwestern Polytechnical University (NPU), Xi'an, China. His research interests include computer vision and pattern recognition, especially focusing on 3D scene reconstruction , compression, understanding, and generation.
\end{IEEEbiography}

\begin{IEEEbiography}[{\includegraphics[width=1in,height=1.25in,clip,keepaspectratio]{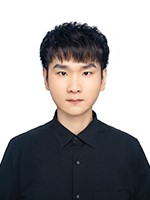}}]{Bohan Li} (Student Member, IEEE) received the B.E. degree from the School of Control Engineering, Northeastern University (NEU),
Shenyang, China, in 2019. He received the M.E. degree from the School of Control Science and Engineering, South China University of Technology
(SCUT), Guangzhou, China, in 2022.
He is currently pursuing the Ph.D. degree in Shanghai Jiao Tong University (SJTU) and Eastern Institute of Technology (EIT). His research interests include 3D visual perception, robotics, and multi-modality content generation.
\end{IEEEbiography}

\vspace{11pt}

\begin{IEEEbiography}[{\includegraphics[width=1in,height=1.25in,clip,keepaspectratio]{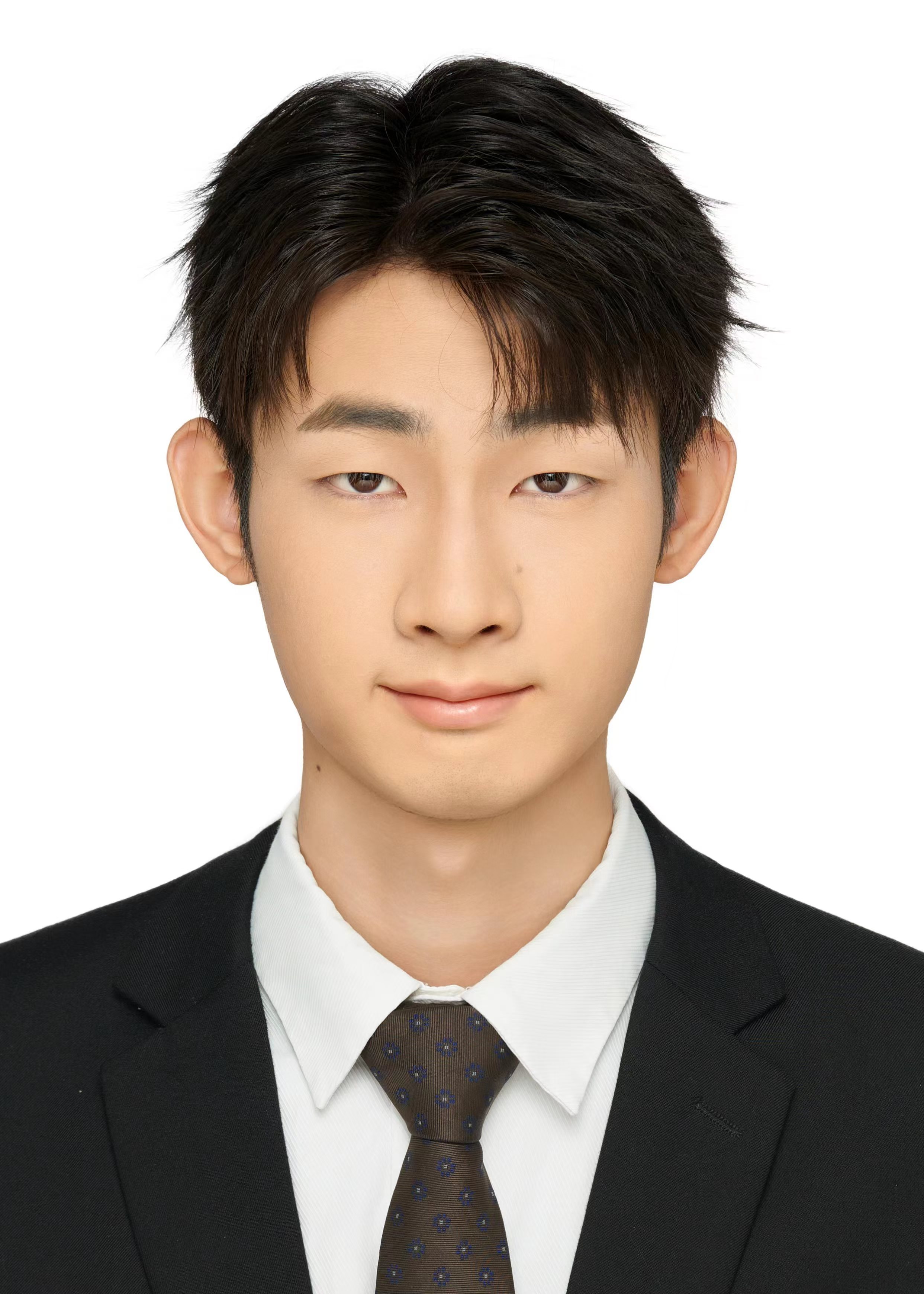}}]{Bingquan Dai}
received the BE degree from the School of Astronautics, Beihang University, China, in 2023. He is currently pursuing a Master's degree at the Shenzhen International Graduate School, Tsinghua University. His research interests include multimodal understanding, 3D reconstruction, and multimodal large models.
\end{IEEEbiography}

\vspace{15pt}

\begin{IEEEbiography}[{\includegraphics[width=1in,height=1.25in,clip,keepaspectratio]{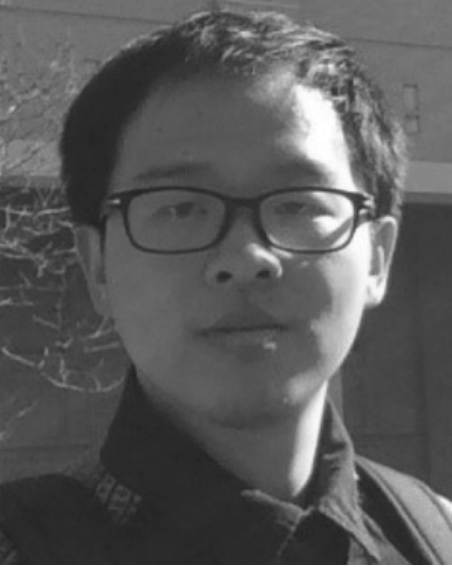}}]{Dingwen Zhang}
is a professor with School of Automation, Northwestern Polytechnical University, Xi'an, China. He received his Ph.D. degree from NPU in 2018. From 2015 to 2017. he was a visiting scholar at the Robotic Institute, Carnegie Mellon University, Pittsburgh, United States. His research interests include computer vision and multimedia processing, especially on saliency detection and weakly supervised learning.
\end{IEEEbiography}

\vspace{18pt}

\begin{IEEEbiography}[{\includegraphics[width=1in,height=1.25in,clip,keepaspectratio]{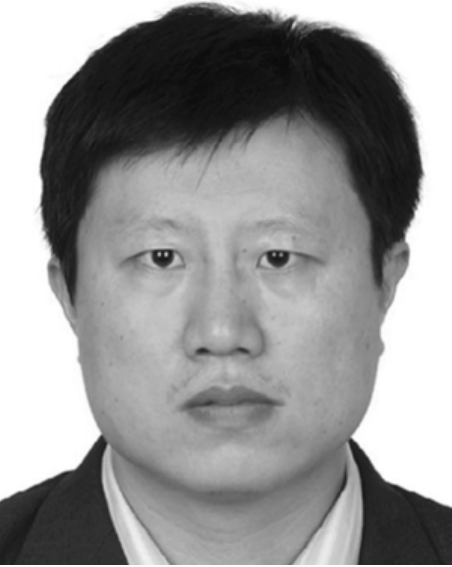}}]{Junwei Han}
(M’12-SM’15) received the PhD degree from Northwestern Polytechnical University, in 2003. He is with the School of Artifical Intelligence, Chongqing University of Posts and Telecommunications. His research interests include computer vision and brain imaging analysis. He has published more than 100 papers in IEEE TRANSACTIONS and top tier conferences. He is a senior member of the IEEE.
\end{IEEEbiography}

\vfill

\end{document}